\documentclass[letterpaper,journal]{IEEEtran}

\usepackage{amsmath,amsfonts,amssymb}
\usepackage{algorithmic}
\usepackage{algorithm}
\usepackage{array}
\usepackage[caption=false,font=normalsize,labelfont=sf,textfont=sf]{subfig}
\usepackage{textcomp}
\usepackage{stfloats}
\usepackage{url}

\usepackage{verbatim}
\usepackage{graphicx}
\usepackage{cite}
\usepackage{siunitx}
\usepackage{booktabs}

\usepackage{fancyhdr}
\fancypagestyle{preprintnotice}{\fancyhf{}%
  \fancyhead[C]{\parbox{\textwidth}{\centering\small\itshape This work has been submitted to the IEEE for possible publication.
  Copyright may be transferred without notice, after which this version may no longer be accessible.}}%
  \fancyfoot[R]{\thepage}}
\begin{document}

\title{Emergent Cooperative Thermoregulation in Lunar Swarm Robotics}

\author{Ana\"is Jeger$^{1,2}$, Harry Holt$^{2}$, and Lennart Puck$^{3}$\\[3pt]
{\normalsize $^{1}$Center for Bits and Atoms, MIT \quad
$^{2}$Advanced Concepts Team, ESA \quad
$^{3}$Planetary Robotics Laboratory, ESA}%
\thanks{Work carried out while A.J. was at ESA; she is now at MIT (anais98@mit.edu).}%
}

\bstctlcite{IEEEexample:BSTcontrol}
\maketitle
\thispagestyle{preprintnotice}

\begin{abstract}
Swarms of small robots are an attractive way to survey lunar caves and lava
tubes. Missions such as NASA's CADRE will fly cooperating rovers for
distributed mapping, but only in daylight. The 354\,h lunar night drops the
surface towards 95\,K, and the radioisotope heaters that let single rovers
survive it cannot be supplied at swarm scale. In this study we investigate
whether a swarm can instead use its own members as thermal shielding.
We train one policy, shared by all twenty robots, by reinforcement learning
with an evolution-strategies estimator. The policy receives a single reward,
once, at the end of the night: the mean temperature of the five coldest
robots. Nothing in that reward refers to position, spacing, exposure or
motion. Two behaviors nonetheless emerge. The swarm aggregates, and robots
circulate between the sheltered core and the exposed rim, the same pair of
behaviors with which emperor penguins survive the Antarctic winter.
We measure the survival threshold, the smallest stored energy per robot at
which every robot is still alive at sunrise. Across ten runs it is
$155.6 \pm 1.3$\,pcm-h, where one pcm-h is 7.2\,kJ of stored heat. That is
$37\,\%$ below a motionless huddle and within $13\,\%$ of a physics-based
lower bound, and at a budget where the motionless huddle loses six robots,
every robot survives. The behavior needs no shared spatial reference and no
robot--robot contact, and transfers from twenty robots to forty without
retraining.
\end{abstract}


\begin{figure}[!t]
\centering
\includegraphics[width=\columnwidth]{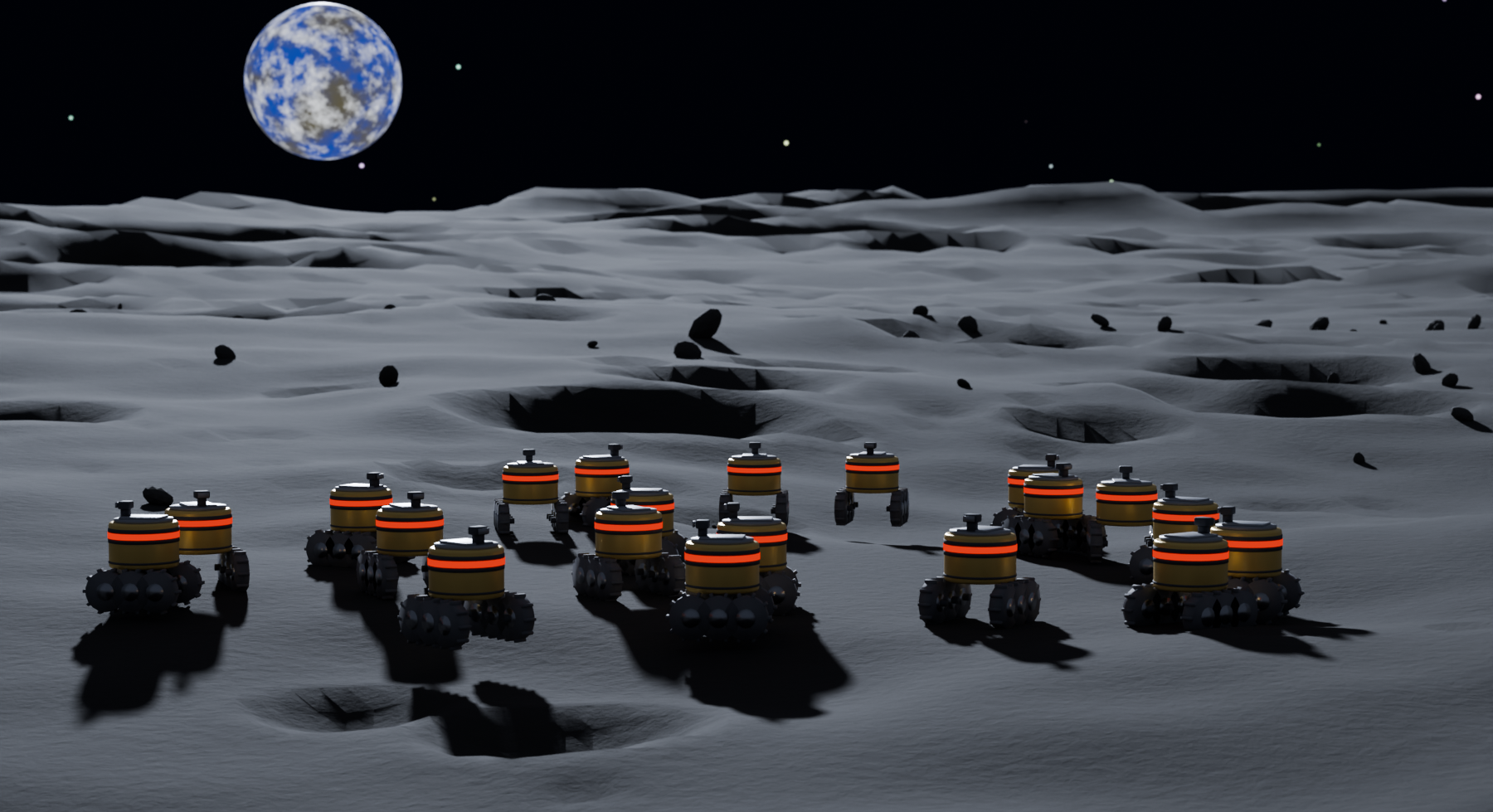}\\[2pt]
\includegraphics[width=\columnwidth]{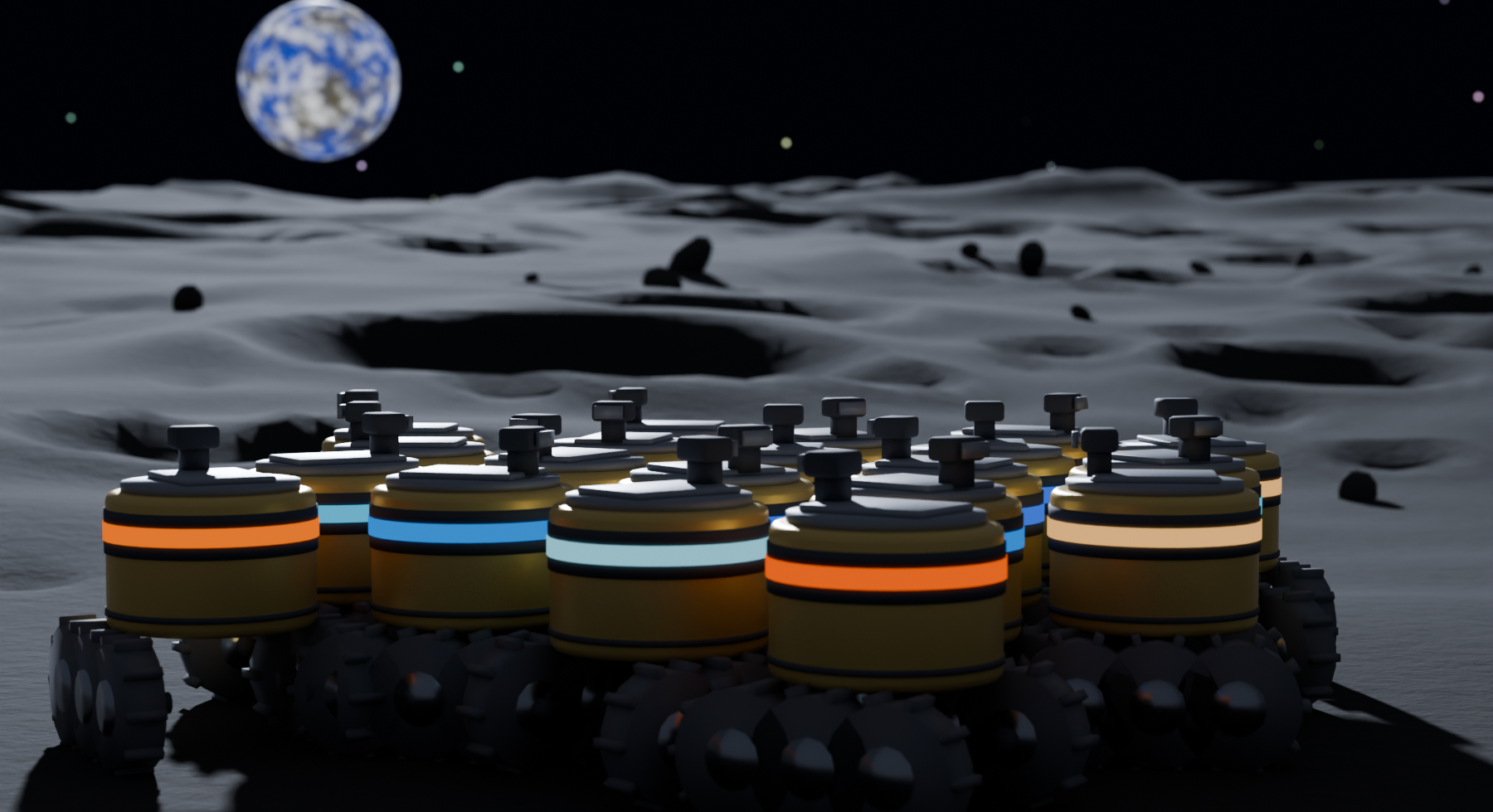}
\caption{Twenty simulated robots during the \SI{354}{\hour} lunar night, from
one rollout of the learned policy. Ring color is each robot's sky exposure
$\chi$ on the scale of Fig.~\ref{fig:emergence}. \textbf{Top:} the scattered
start at hour 0. Every robot is red: with no neighbor close enough to block it,
each one radiates to the full sky. \textbf{Bottom:} hour 73, after the swarm
has gathered. The interior is blue, shielded by the bodies around it, and only
the rim stays warm-colored; all twenty are alive.}
\label{fig:teaser}
\end{figure}

\section{Introduction}
Surviving the lunar night as a small rover remains a difficult problem due to temperatures falling towards \SI{95}{\kelvin} and its duration of approximately \SI{354}{\hour} \cite{williams2017,nasaMoonlight}. Recently, swarm robotics has become increasingly popular for mapping extraterrestrial environments \cite{cadre,thangavelautham2017,kalita2018}. This is especially useful for exploring underexplored environments such as lunar caves and lava pits, where the loss of one small robot would not necessarily compromise the entire mission \cite{thangavelautham2017,kalita2018}. NASA's CADRE mission, for example, will use three small cooperating rovers for distributed lunar mapping, but the rovers are only designed to operate during the daylight hours of a single lunar day and will therefore not survive the lunar night \cite{cadre}.

While some rovers have managed to withstand the extreme temperatures of the lunar night, the solutions used to achieve this are difficult to transfer to small swarm robots. Yutu-2, for example, uses radioisotope-assisted thermal control and has successfully resumed operation after lunar nights \cite{biswas2021,cnsa2019night}. Pragyan, in contrast, was designed for one lunar day and did not resume communication after the lunar night \cite{isroCh3Details,isroPragyanNoSignal}. Radioisotope Heating Units (RHUs) have allowed lunar systems to survive the night, but their fuel is scarce: holding a \SI{2}{\watt} heater needs about \SI{5}{\gram} of plutonium dioxide per robot, two standard units' worth \cite{zillmer2022lwrhu}, and the US production target of roughly \SI{1.5}{\kilogram} per year \cite{doePu238} is sized for flagship radioisotope power systems rather than for expendable robots.

So before focus shifts towards sophisticated autonomy or science objectives such as mapping, a lunar robot swarm first needs to be capable of surviving the lunar night. In this study we investigate whether a swarm of robots can use each other as thermal shielding, and so reduce the stored energy each robot needs to survive the lunar night. We take the approach from emperor penguins, which huddle to reduce heat loss while continuously exchanging positions between the exposed outside and the sheltered core \cite{ancel1997,gerum2013,gilbert2006energy,gilbert2006weather,waters2012,zitterbart2011}. We investigate whether a swarm of lunar robots can use this same principle to redistribute exposure between robots and reduce the thermal burden on each individual robot.

We propose a decentralized policy that commands each robot's motion throughout the night from purely local observations, with one parameter set shared across the swarm, and train it by reinforcement learning.
Our contributions are: (i) we train a shared policy by reinforcement learning on a purely thermodynamic objective (ii) both aggregation and sustained core–rim circulation emerge from it, and together they keep every robot alive at a fixed energy budget where a motionless huddle does not (iii) the behavior requires no shared spatial reference and (iv) it generalizes beyond its training swarm size, controlling swarms from half to twice that number without retraining.

\section{Related Work}

While there have been works related to thermoregulation in robots, such as BEECLUST-style swarm controllers which were used to aggregate based on favorable light or temperature conditions, this was mostly done using local sensing and waiting/motion rules based on the environmental field \cite{schmickl2009,kernbach2009,schmickl2011beeclust}. In the same vein, gradient-based aggregation and large-scale self-assembly have been demonstrated with simple robots such as Kilobots \cite{rubenstein2014}. The key difference in our paper is that the robots themselves have to create a thermally favorable configuration by shielding each other instead of moving to a favorable external location.

Previous work has been done on penguin-inspired multi-robot huddling. Mina and Min proposed tightly packed formations in which robots share energy through inductive coupling while shuffling positions, as well as a boundary-movement strategy in which exposed robots relocate towards a more sheltered side of the group \cite{mina2018ssrr,mina2018robio}. These approaches already show the usefulness of penguin-inspired position
exchange for group survival. Neither work models a lunar night, however, and
neither treats the robots' own bodies as radiative shielding; their mechanisms
are inductive energy sharing and boundary movement in generic cold conditions. Earlier work \cite{jeger2026cooperative} first proposed using a robot swarm's own members as thermal shielding to survive the lunar night, showing in ARGoS that a shared Proximal Policy Optimization policy with a phase-conditioned reward and a discrete huddle primitive reached $55\%$ survivors over a compressed lunar cycle, with the perimeter lost. The present paper removes the phase labels, the primitive, and every geometric reward term, extends to the full night, and reaches 20/20 survival.

Previously, evolution strategies (ES) have already been used for long-horizon objectives with delayed or sparse rewards \cite{salimans2017,wierstra2014}, including sparse team-based objectives in multi-agent coordination \cite{khadka2020}. While shared-parameter policies are a standard way to get homogeneous multi-agent behavior \cite{gupta2017,huttenrauch2019}, we introduce a thermodynamic objective that contains no explicit geometric term. Concretely, this implies that the policy is not directly rewarded for huddling, circulation, or equalizing exposure. Instead, the circulation emerges because it helps the swarm survive.

While rovers have previously survived the lunar night, lunar-night survival is
generally treated as a single-robot problem. Current methodologies mostly focus
on how much stored energy an individual robot needs or how much radioisotope
heating it requires \cite{biswas2021}; design tools for the sub-\SI{50}{\kilogram}
class couple terramechanics, mass, power and thermal survival into a per-vehicle
evaluator \cite{roverdevkit2026}. Heater-free concepts do exist, but often rely
on the robot returning to a shelter at the lander \cite{lee2019}. In our work, we
focus on how the swarm itself can reduce each robot's thermal burden through
collective shielding and alternating exposure.

\section{Simulation and Thermal Model}

We simulate $N=20$ robots over a full \SI{354}{\hour} lunar night and discretize it into 23,556 time steps. Here, every thermal step corresponds to \SI{54.1}{\second}. The stored energy is given in kJ. When we compare survival thresholds, we also use pcm-h, where $1\,\mathrm{pcm\text{-}h}=\SI{7.2}{\kilo\joule}$ for the \SI{2}{\watt} heater.

\subsection{Thermal Dynamics}

We model radiative loss as linear in temperature about the setpoint, which is exact while the heater holds $T_{\mathrm{set}}$ and overestimates loss during
the post-exhaustion coast. The rate is set by the robot's angular sky exposure
$\chi\in[0,1]$:
\begin{equation}
\dot{Q}_i = G_{\mathrm{sky}}\left[f_{\mathrm{sky}} + (1-f_{\mathrm{sky}})\chi_i\right](T_i-T_{\mathrm{env}}).
\end{equation}
with $G_{\mathrm{sky}}=\SI{0.0108}{\watt\per\kelvin}$, $f_{\mathrm{sky}}=0.275$, and $T_{\mathrm{env}}=\SI{95}{\kelvin}$ \cite{williams2017}. Here, $\chi=0$ represents a fully surrounded robot, while $\chi=1$ represents an isolated robot. The robot maintains $T_{\mathrm{set}}=\SI{273.15}{\kelvin}$ while stored energy remains, and only once it is exhausted will it begin cooling according to its thermal model,
\begin{equation}
\dot{T}=-\frac{\dot{Q}}{C_{\mathrm{th}}},
\end{equation}
where $C_{\mathrm{th}}=\SI{2000}{\joule\per\kelvin}$. The robot is considered dead once its temperature falls below $T_{\mathrm{death}}=\SI{233.15}{\kelvin}$.
The heater can hold the setpoint even in the worst case: a fully isolated robot
loses \SI{1.92}{\watt} against a \SI{2.0}{\watt} heater, and with
$C_{\mathrm{th}}=\SI{2000}{\joule\per\kelvin}$ a \SI{0.1}{\watt} shortfall
sustained over one timestep would move the temperature by \SI{3}{\milli\kelvin}.

Integrating (1) and (2) over the night gives the stored energy a robot needs as
a function of its time-averaged exposure alone, because the loss coefficient is
affine in $\langle\chi\rangle$. While the heater is active the store pays for
the loss required to hold $T_{\mathrm{set}}$; once it is exhausted the robot
coasts from $T_{\mathrm{set}}$ to $T_{\mathrm{death}}$, so that final part of
the night need not be covered by stored energy.

\begin{equation}
E^*(\langle\chi\rangle)
=
k(\langle\chi\rangle)\Delta T\,t_{\mathrm{night}}
-
\Lambda,
\end{equation}
where
\begin{equation}
k(\langle\chi\rangle)
=
G_{\mathrm{sky}}
\left[
f_{\mathrm{sky}}
+
(1-f_{\mathrm{sky}})\langle\chi\rangle
\right],
\end{equation}
and
\begin{equation}
\Lambda
=
C_{\mathrm{th}}\Delta T
\ln
\left(
\frac{\Delta T}
{T_{\mathrm{death}}-T_{\mathrm{env}}}
\right),
\qquad
\Delta T=T_{\mathrm{set}}-T_{\mathrm{env}}.
\end{equation}

Evaluating this expression at the mean exposure of the densest packing gives the $N=20$ equalization bound of $138.1\,\mathrm{pcm\text{-}h}$. We use this value as a physics-based lower bound against which all survival thresholds are normalized.
\label{sec:bound}

\subsection{Body Geometry}

We do not assume a robot geometry. Instead we invert the radiative balance at
the setpoint to constrain the product $\varepsilon A$, where $\varepsilon$ is
the emissivity and $A$ the radiating area, which leaves a family of thermally
consistent designs rather than a single one.

The dynamics use a body mass of $M=2.0$\,kg; the thermal mass
$C_\mathrm{th}=2000$\,J\,K$^{-1}$ corresponds to approximately 2.1\,kg of
aluminum-equivalent structure and avionics. We compare this to small mobile robots with indicative bulk densities of roughly \SI{200}{\kilogram\per\metre\cubed} to \SI{700}{\kilogram\per\metre\cubed}, based on published dimensions and masses of platforms such as the TurtleBot3 Burger, Khepera IV, and e-puck \cite{robotisTurtlebot3,soaresKhepera4,gctronicEpuck}.
In our simulations, we use $\varepsilon=0.035$, which gives $A=\SI{0.177}{\metre\squared}$, $H=\SI{0.246}{\metre}$, and $\rho=\SI{377}{\kilogram\per\metre\cubed}$. This emissivity also happens to correspond to the ten-layer blanket flown on the Chang'e-4 LND instrument, although this coincidence is not necessary for the results presented here \cite{wimmer2020}.

The locomotion and contact parameters of Table~\ref{tab:params} are operating
points with measured failures on either side, not free choices. The commanded
speed fixes the largest force one robot can exert through
$F=\mu v_{\mathrm{cmd}}$; below about \SI{50}{\milli\newton} the swarm still
drives the same total distance but stops penetrating the pack, and exposure
returns almost to its static value. Contact compliance is constrained in the
same way: the yield transition that lets a robot burrow through its neighbors
exists only for compliant contact, and at \SI{600}{\newton\per\metre} force
transmits instantly across the whole cluster, leaving no yield stress to exceed
and collapsing the behavior back to a jammed huddle. The substep count then
follows from the contact stiffness, since explicit integration is stable for
$\mathrm{SUBD}>2K_{\mathrm{con}}\Delta t/\mu$, which is 109 at these values.

\begin{table}[t]
\centering
\small
\caption{Model parameters.}
\label{tab:params}
\begin{tabular}{@{}llll@{}}
\toprule
Parameter & Value & Parameter & Value \\
\midrule
$N$ & 20 & $T_{\mathrm{set}}$ & \SI{273.15}{\kelvin} \\
$R$ & \SI{0.085}{\metre} & $T_{\mathrm{env}}$ & \SI{95}{\kelvin} \\
$M$ & \SI{2.0}{\kilogram} & $T_{\mathrm{death}}$ & \SI{233.15}{\kelvin} \\
$t_{\mathrm{night}}$ & \SI{354}{\hour} & $C_{\mathrm{th}}$ & \SI{2000}{\joule\per\kelvin} \\
$\Delta t$ & \SI{54.1}{\second} & $G_{\mathrm{sky}}$ & \SI{0.0108}{\watt\per\kelvin} \\
$v_{\mathrm{cmd}}$ & \SI{0.015}{\metre\per\second} & $Q_{\max}$ & \SI{2.0}{\watt} \\
$\mu$ & \SI{6}{\newton\second\per\metre} & $f$ & 0.275 \\
$K_{\mathrm{con}}$ & \SI{6}{\newton\per\metre} & $\varepsilon$ & 0.035 \\
$\mu_s$ & 0.5 & $P_{\mathrm{elec}}$ & \SI{0.5}{\watt} \\
\bottomrule
\end{tabular}
\end{table}

\subsection{Exposure}

We compute sky exposure using angular occlusion: each robot's horizon is divided into 36 parts, and each part is considered shielded if a neighbor within $3R$ subtends it. The neighbor's effective subtended half-angle is
\begin{equation}
1.6\arctan(R/d),
\end{equation}
where $d$ is the center-to-center distance. The factor 1.6 is fitted: we scan it against the Monte Carlo view-factor calculations and take the value giving the best affine fit between the proxy and the true loss
($R^2=0.97$, $3.7\,\%$ RMS). It exceeds unity because a neighboring cylinder blocks both its own angular width and some of the low-elevation sky behind it, which a two-dimensional horizon slice does not capture. A living robot provides
full shielding; one that has cooled towards $T_{\mathrm{env}}$ is treated as a partial shield. $\chi$ is therefore determined entirely by the current geometry of the swarm, and is the only coupling between locomotion and the thermal
state.

\section{Method}
\subsection{Policy}

All robots share one completely decentralized policy, a recurrent network with
a 64-unit hidden state. We use a recurrent policy because a robot's
exposure history is not recoverable from a single observation;
Sec.~\ref{sec:memory} then asks whether that memory is actually used.
Each robot has access to its own temperature $T$, its
stored-energy fraction, its current sky exposure $\chi$, the number of contacts,
the unit vector and range to the contact centroid, the mean and minimum neighbor
temperature and stored energy, and a unit vector and range to a shared spatial
reference. This gives fourteen observation channels in total. Everything is
local, and the robots do not have a global picture of the swarm. The same
learned parameters $\theta$ are deployed across the whole swarm.

\begin{figure*}[t]
\centering\includegraphics[width=\textwidth]{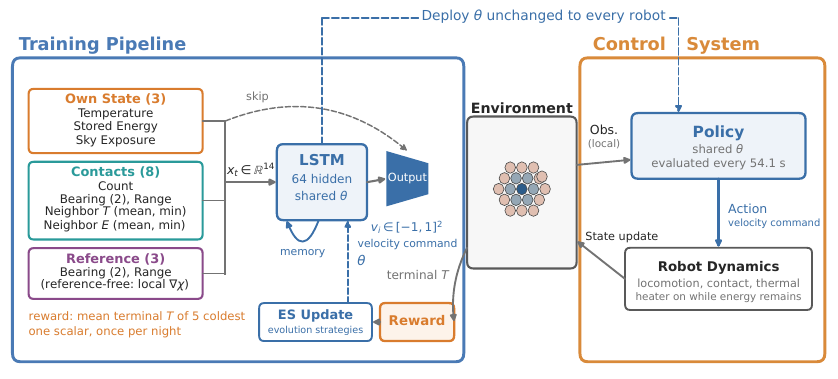}
\caption{Overview of the learning system. \textbf{Left:} fourteen local
observation channels enter a 64-unit LSTM whose state persists across thermal
steps; a skip connection passes the input directly to the output layer, which
produces a velocity command. ES perturbs the shared
parameters $\theta$ and updates them from a single scalar returned after the
full night.
\textbf{Middle:} the simulated swarm, colored by sky exposure $\chi$.
\textbf{Right:} the control loop deployed on every robot, using the trained
$\theta$ unchanged.}
\label{fig:system}
\end{figure*}

Fig.~\ref{fig:system} shows the full system.

The optimization term consists of the terminal temperature of the coldest quintile of robots. For $N=20$, this corresponds to $K=5$. We use the coldest quintile rather than the swarm average so that the objective is dominated by the robots closest to failure, while remaining less sensitive to a single outlier. We use a sparse reward which is returned only once, at the end of the lunar night, and no reward shaping takes place during the episode. We leave out any terms which could influence our emergence claim, such as position, spacing, sky exposure, circulation, huddling, or motion. The only motivating factor is therefore to improve survival.

\begin{equation}
R(\pi)=\frac{1}{K}\sum_{i\in\mathcal C_K}T_i(t_{\mathrm{night}}),
\qquad K=5.
\end{equation}

\subsection{Training}

Our optimization routine consists of using reinforcement learning, implemented directly in JAX. We use the antithetic, rank-normalized ES estimator of Salimans et al.~\cite{salimans2017}, with a population of 64, $\sigma=0.02$, and a learning rate of $0.005$. Each generation samples paired perturbations $\epsilon$ and $-\epsilon$, replaces the raw fitness values by centered ranks, and updates the shared policy parameters $\theta$ using the resulting fitness-weighted perturbations. Every candidate is evaluated over 8 starts, and training runs for 200 generations. After 25 generations without improvement the parameters are rolled back to the best set seen so far and the run continues to 200. The population and evaluation count are not free: $64\times8$ is the smallest setting at which training reliably finds the survival basin, and below it we observe a specific failure, in which the number of living robots decays to zero while the objective continues to improve.

We use ES because the reward is a single scalar returned only after the full \SI{354}{\hour} episode, with no intermediate signal from which to estimate per-step advantages over 23,556 decisions. In addition, the objective is a swarm-level quantity, given by the mean terminal temperature of the coldest robots, rather than a per-agent return.

The scale of the training budget comes from the physics, but the value
does not. We express the budget as a multiple of the equalization bound of
Sec.~\ref{sec:bound}, which fixes what ``enough stored energy'' means
independently of any controller; the multiple itself is a design choice, and we
select it empirically (Table~\ref{tab:budget}). We train at $1.18\times$ the
bound. At that budget 14/20 robots survive in a motionless huddle and 0/20
survive if dispersed: high enough that survival is reachable, low enough that
huddling alone does not achieve it. The lower edge of that window follows from
the reward itself. Below roughly $1.05\times$ the bound no policy keeps all
twenty alive, so the terminal temperature of the coldest five only ranks how
far the dead robots cooled, and the optimizer has no path towards survival to
follow. We train under four conditions, which differ only in the initial
configuration (packed or scattered) and in whether the shared-reference channels
are present; the architecture, reward, budget and optimizer settings are
identical in all four. We run several independent training runs per condition;
Table~\ref{tab:main} gives the number pooled for each.

Because we train on temperature and never optimize stored energy directly, the
survival threshold measures the policy that training happens to produce rather
than the training objective itself, and is therefore sensitive to the budget
(Table~\ref{tab:budget}). Training at $1.40\times$ the bound leaves a motionless
huddle keeping 19 of 20 robots, so there is little left to learn and the
resulting policy reaches only \SI{176.4}{} pcm-h. Below $1.18\times$ the
threshold is flat: $1.10\times$ gives $155.8$ against $155.7$ at $1.18\times$,
and the policy trained at $1.10\times$ ends with a threshold above its own
training budget. We therefore use $1.18\times$, the largest budget that still yields the best threshold.

\begin{table}[t]
\centering\small
\caption{Sensitivity of the survival threshold to the training budget. The
budget is an input, the threshold an outcome; both are in pcm-h. One policy per
budget, same seed.}
\label{tab:budget}
\begin{tabular}{@{}rrrr@{}}
\toprule
Training budget & Training budget & Resulting threshold & Static \\
($\times$ bound) & (pcm-h) & (pcm-h) & survivors \\
\midrule
1.10 & 151.9 & 155.8 & 14/20 \\
1.18 & 163.0 & 155.7 & 14/20 \\
1.40 & 193.3 & 176.4 & 19/20 \\
\bottomrule
\end{tabular}
\end{table}

\subsection{Evaluation}

Our policy's survival threshold is the smallest initial stored-energy budget for which all robots survive. We require all robots to survive because a policy which only preserves the average swarm state can still allow individual robots to fail. This condition has to hold across all ten held-out evaluation seeds. We find the threshold using bisection and continue until we reach $0.5\,\%$ precision. The evaluation seeds are held out from training: each seed draws a different initial configuration, so the thresholds measure every policy on starting configurations it never saw while training.

\section{Results}

\subsection{Circulation Emerges}
\label{sec:emergence}

\begin{figure}[t]
\centering
\includegraphics[width=\columnwidth]{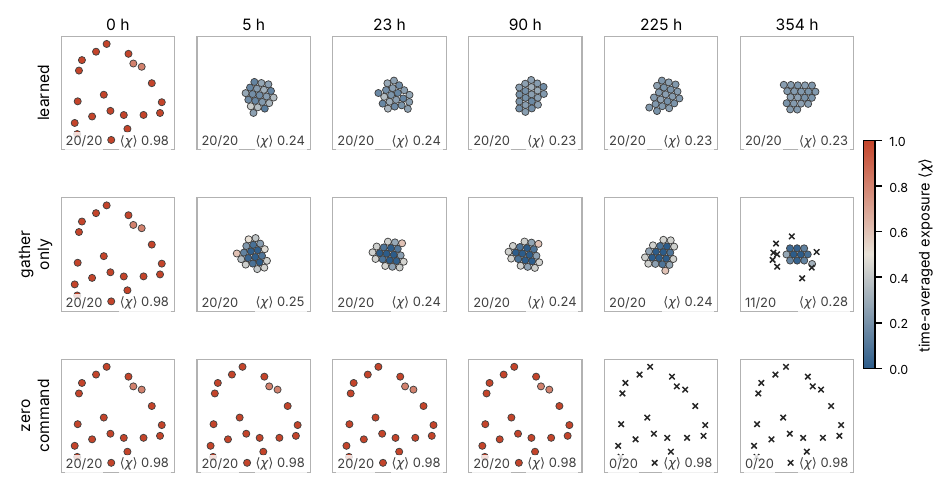}
\caption{Emergence from a scattered start; all rows share one initial
configuration. Color is each robot's time-averaged exposure
$\langle\chi\rangle$ up to that frame; crosses are dead robots. Under each
frame: robots alive and the swarm's mean $\langle\chi\rangle$. The thresholds
of Table~\ref{tab:main} are a separate quantity, bisected on the budget rather
than read off these runs.
\textbf{Top:} the learned policy gathers within five hours and circulates,
every average converges, 20/20 survive. \textbf{Middle:} the gather-only
rule, which aggregates but never circulates, forms the same huddle, but a
core--rim gradient persists all night and nine robots are lost. \textbf{Bottom:} under
zero command the swarm stays scattered and all are lost by 225\,h.}
\label{fig:emergence}
\end{figure}

Training from a packed start begins at $14/20$ survivors, which is the
motionless-huddle result: the output head is initialized to zero, so the
generation-0 policy commands no motion at all. During training, we reach $20/20$ survival within tens of
generations. As a by-product, the worst robot's time-averaged exposure drops
from about 0.71 to 0.26, close to the swarm mean of 0.23, even though exposure
is not included as a term in the reward. From a scattered start, the swarm
first aggregates and then circulates (Fig.~\ref{fig:emergence}). We compare
these two configurations against two baselines. The first is a zero-command
baseline, which remains scattered and in which all robots die. The second is a
\emph{gather-only} rule: a hand-coded controller in which every robot leans
gently inward all night and nothing else. No term ever drives a robot back
outward, so the swarm aggregates and then holds its arrangement, which isolates
aggregation from circulation. It steers on the shared reference, so it is not
reference-free. Taking these test cases together, we arrive at a claim of emergence: the policy is rewarded only on the terminal temperature of its coldest members and not on gathering, huddling, exposure, or circulation.

\begin{table}[t]
\centering
\small
\caption{Survival thresholds in pcm-h. Learned entries pool converged runs.}
\label{tab:main}
\begin{tabular}{@{}lrrr@{}}
\toprule
Condition & $n$ & Threshold & $\times$ bound \\
\midrule
Dispersed, no command      & --- & 328.5 & 2.38 \\
Motionless huddle          & --- & 246.0 & 1.78 \\
Gather-only rule           & --- & 250.3 & 1.81 \\
\midrule
Learned, packed, shared ref.       & 3 & $155.7 \pm 1.3$ & 1.13 \\
Learned, packed, no shared ref.    & 2 & $154.3 \pm 1.6$ & 1.12 \\
Learned, scattered, shared ref.    & 2 & $155.4 \pm 0.6$ & 1.13 \\
Learned, scattered, no shared ref. & 3 & $156.7 \pm 1.1$ & 1.13 \\
\midrule
All converged runs         & 10 & $155.6 \pm 1.3$ & 1.13 \\
Hand-coded rule            & --- & 154.5 & 1.12 \\
\midrule
Equalization bound         & --- & 138.1 & 1.00 \\
\bottomrule
\end{tabular}
\end{table}

Across all converged learned runs the survival threshold is
$155.6 \pm 1.3\,\mathrm{pcm\text{-}h}$ (Table~\ref{tab:main}): $37\,\%$ below
the motionless huddle, $53\,\%$ below the dispersed baseline, and $12.7\,\%$
above the equalization bound. With
two to three runs per condition the four learned conditions are all within the
run-to-run spread, so we claim no ordering among them; the pooled value is the
one to read.

We also evaluated each condition across a range of budgets. At a common budget
of \SI{150}{} pcm-h, below every condition's threshold, the coldest-five
terminal temperature reads \SI{99}{\kelvin} for a dispersed swarm,
\SI{127}{\kelvin} for the gather-only rule, \SI{147}{\kelvin} for a motionless
huddle and \SI{232}{\kelvin} for a circulating one.
Above each condition's threshold the reward
saturates at $T_{\mathrm{set}}$ by construction, since the heater holds the
setpoint while stored energy remains, which is why we compare thresholds rather
than terminal temperatures. The gather-only rule reaches $20/20$ only
at \SI{250.3}{} pcm-h, essentially where a motionless huddle does at $246$:
forming the huddle is not by itself sufficient, and circulation carries the
requirement down to $155$.

The hand-coded rule performs nearly identically to the learned policies,
reaching $154.5\,\mathrm{pcm\text{-}h}$. The rule is simple: every robot leans
gently inward throughout the night, and whichever robot holds the least stored
energy among its contacts burrows inward at full force until it reaches the
core. The gather-only baseline of Sec.~\ref{sec:emergence} is this same rule
with the burrow term removed.

\subsection{No Shared Reference Is Required}
\label{sec:nobeacon}

A shared reference is a natural feature for a swarm that must reconvene, but it
is a global observation, and aggregation towards it could pass for simple
navigation rather than decentralized coordination. We therefore remove it from
training, leaving no external frame or shared infrastructure.

Performance remains essentially unchanged: the packed-start threshold changes
from $155.7 \pm 1.3\,\mathrm{pcm\text{-}h}$ with a shared reference to
$154.3 \pm 1.6\,\mathrm{pcm\text{-}h}$ without one, while the scattered-start
threshold changes from $155.4 \pm 0.6\,\mathrm{pcm\text{-}h}$ to
$156.7 \pm 1.1\,\mathrm{pcm\text{-}h}$. Just like the policy with a reference,
the reference-free policy aggregates, circulates, and survives at approximately
the same cost. However, the robots move more, with roughly 4500 crossings
compared with 1800 with the shared reference, while achieving similar exposure.

\subsection{The Role of Memory}
\label{sec:memory}
Memory is not required, but it is not unused either. Feed-forward policies with
the same behavioral parameter count reach
$160.8 \pm 2.8\,\mathrm{pcm\text{-}h}$. We also test each trained
recurrent policy with its hidden state zeroed at every step, which removes the
recurrence and reduces the network to a feed-forward map from the current
observation to the command; it asks whether a policy uses its memory at
deployment. Of the ten converged recurrent
policies, five reach their threshold unchanged with the hidden state zeroed at
every step, having never learned to use it, while the other five collapse to
$190$--$316\,\mathrm{pcm\text{-}h}$ without it. The same architecture,
objective, and threshold therefore admit both memoryless and memory-dependent
solutions. Recurrence is therefore not required for the behavior: a
feed-forward policy costs about $3\,\%$ in survival threshold, and half the
recurrent policies never use their hidden state at all. Qualitatively similar circulation was also observed in the earlier ARGoS implementation \cite{jeger2026cooperative}, which uses frictional, non-interpenetrating contact and a different reward.

\subsection{Transfer to Other Swarm Sizes}
\label{sec:scaling}

\begin{figure}[t]
\centering\includegraphics[width=\columnwidth]{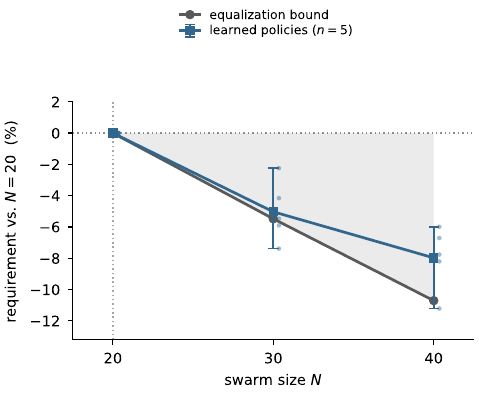}
\caption{Transfer to larger swarms. Change in survival requirement relative to
$N=20$, the size at which all five policies were trained; they were evaluated at
$N=30$ and $N=40$ without retraining. Squares are the mean across policies with
bars spanning best and worst; faint points are the individual policies. The grey
region is the change available from the equalization bound.}
\label{fig:scaling}
\end{figure}

Five of the converged policies, the first five by seed index and taken
without reference to their thresholds, were evaluated at $N=20,30,40$ without retraining, and
the survival requirement decreased as the swarm size increased. Relative to
$N=20$, the requirement falls by $5.0\,\%$ on average at $N=30$ and $8.0\,\%$ at
$N=40$, against a bound that falls by $5.5\,\%$ and $10.7\,\%$
(Fig.~\ref{fig:scaling}). The policies therefore track the geometric benefit of
a larger swarm without retraining, capturing most of what the bound makes
available at $N=30$ and about three quarters of it at $N=40$.

\subsection{What the Behavior Depends On}
\label{sec:ablation}

At evaluation, one observation channel is zeroed at a time. Each policy was
trained with all fourteen channels, so this measures robustness to losing one
at deployment rather than dependence during learning, and the policies are not
equally robust. The three policies s0--s2 are the first three by seed
index, again taken without reference to their thresholds. Neighbor stored
energy is the only channel which is consistently cheap to remove, increasing
the survival threshold by only $0.3$--$2.7\,\%$. In contrast, neighbor
temperature is important for one policy but has almost no effect on the others,
while removing the robot's own sky exposure has little effect on two policies
but causes a large degradation in the third. This shows that similar
macroscopic behavior and survival thresholds can emerge from different
dependencies on local information. We therefore report the ablation results
per policy rather than using a pooled mean, which would hide these differences.
Zeroing both neighbor channels is the communication-free case: the robot
still senses that it has contacts, and their bearing and range, but receives no
telemetry from them. One policy is indifferent to this ($154.9$), one degrades
($175.5$), and one fails ($268.4$), so a purely self-centered observation is
sufficient for some solutions and not for others.

\begin{table}[t]
\centering
\small
\caption{Observation ablation results, reported per policy.}
\label{tab:ablation}
\begin{tabular}{@{}lrrr@{}}
\toprule
Channel zeroed & s0 & s1 & s2 \\
\midrule
None                    & 157.1 & 155.8 & 155.4 \\
Neighbor stored energy  & 161.4 & 157.5 & 155.8 \\
Neighbor temperature    & \textbf{179.8} & 154.9 & 159.6 \\
Own sky exposure        & 157.1 & 159.2 & \textbf{242.6} \\
Both neighbor channels  & 175.5 & 154.9 & \textbf{268.4} \\
\bottomrule
\end{tabular}
\end{table}

\section{Limitations}

The policies were trained with physical contact. However, due to the abrasion
that repeated contact could cause, we transferred the policy to a
\SI{25}{\milli\metre} keep-out distance enforced as a velocity constraint. The
same policies survive without retraining with separation maintained all night,
so direct contact is not required, but preventing it costs about $13\,\%$ in
stored energy. This extra cost is independent of the
controller and comes from the reduced shielding caused by separation rather
than from a failure of the learned policy.

While robot--robot contact can be avoided, thus minimizing abrasion between the bodies, the robots are in motion throughout the night. We log an average travel of \SI{4.3}{\metre\per\hour} per robot across the full \SI{354}{\hour} night, against a commanded \SI{15}{\milli\metre\per\second} (\SI{54}{\metre\per\hour}): the robots are stationary for most of the night and move in short repositioning bursts. Net displacement over the night is under \SI{1}{\metre}, so the travel is circulation rather than drift. Low speeds notwithstanding, this is sustained wheel--regolith interaction over 354 hours, and abrasion therefore remains a concern. The current model does not capture wear or the resulting drift in contact or friction properties, so long-duration mechanical durability has not yet been validated. Future work will involve transferring the policies to hardware and testing them under realistic regolith and contact conditions.

\section{Conclusions}

A swarm can reduce its lunar-night thermal burden by using its own members as
radiative shielding. Rewarded only on the terminal temperature of its five
coldest robots, the policy aggregates and circulates between core and rim with
no term for huddling, exposure, or motion, and reaches a survival threshold of
$155.6 \pm 1.3\,\mathrm{pcm\text{-}h}$: $37\,\%$ below a motionless huddle
and $12.7\,\%$ above the equalization bound. The behavior needs no shared
reference, recurrence, or robot--robot contact, and transfers from $N=20$ to
$N=40$ without retraining. What remains is physical validation, above all
wheel--regolith wear over a full night.
\IEEEtriggeratref{19}
\bibliographystyle{IEEEtran}
\bibliography{references}

\end{document}